\documentclass[conference]{IEEEtran}
\IEEEoverridecommandlockouts

\usepackage{cite}
\usepackage{amsmath,amssymb,amsfonts}
\usepackage{algorithmic}
\usepackage{graphicx}
\usepackage{multirow}
\usepackage{booktabs}
\usepackage{textcomp}
\usepackage{xcolor}
\usepackage{tikz,lipsum,lmodern}
\usepackage[most]{tcolorbox}

\makeatletter 
\newcommand{\linebreakand}{%
  \end{@IEEEauthorhalign}
  \mbox{}\par
  \mbox{}
  \begin{@IEEEauthorhalign}
}
\makeatother 

\begin{document}

\title{Building AI Service Repositories for On-Demand Service Orchestration in 6G AI-RAN \\
\thanks{The authors are with the Faculty of Engineering and Applied Sciences, Cranfield University, United Kingdom. The work is supported by EPSRC CHEDDAR: Communications Hub for Empowering Distributed clouD computing Applications and Research (EP/X040518/1) (EP/Y037421/1).

Corresponding Author: Yun Tang.}
}

\author{
\IEEEauthorblockN{Yun Tang}
\IEEEauthorblockA{yun.tang@cranfield.ac.uk}
\and
\IEEEauthorblockN{Mengbang Zou}
\IEEEauthorblockA{M.Zou@cranfield.ac.uk} \\
\and
\IEEEauthorblockN{Udhaya Chandhar Srinivasan}
\IEEEauthorblockA{u.srinivasan@cranfield.ac.uk}
\and
\IEEEauthorblockN{Obumneme Umealor}
\IEEEauthorblockA{obumneme.umealor@cranfield.ac.uk}
\linebreakand
\and
\IEEEauthorblockN{Dennis Kevogo}
\IEEEauthorblockA{Dennis.Kevogo@cranfield.ac.uk}
\and
\IEEEauthorblockN{Benjamin James Scott}
\IEEEauthorblockA{benjamin.scott@cranfield.ac.uk}
\and
\IEEEauthorblockN{Weisi Guo}
\IEEEauthorblockA{weisi.guo@cranfield.ac.uk} \\
}

\maketitle

\begin{abstract}
Efficient orchestration of AI services in 6G AI-RAN requires well-structured, ready-to-deploy AI service repositories combined with orchestration methods adaptive to diverse runtime contexts across radio access, edge, and cloud layers. Current literature lacks comprehensive frameworks for constructing such repositories and generally overlooks key practical orchestration factors. This paper systematically identifies and categorizes critical attributes influencing AI service orchestration in 6G networks and introduces an open-source, LLM-assisted toolchain that automates service packaging, deployment, and runtime profiling. We validate the proposed toolchain through the Cranfield AI Service repository case study, demonstrating significant automation benefits, reduced manual coding efforts, and the necessity of infrastructure-specific profiling, paving the way for more practical orchestration frameworks.
\end{abstract}

\begin{IEEEkeywords}
AI Service Provisioning, 6G, AI-RAN
\end{IEEEkeywords}

\section{Introduction}

The significance of AI in 6G RAN is underscored by its potential to revolutionize service delivery and network management. AI-RAN aims to support a wide array of applications, from autonomous vehicles to smart cities, by providing on-demand, context-aware AI services that adapt to user needs in real-time. However, we identify significant bottlenecks towards effective orchestration of on-demand AI services in real-world AI-RAN settings. Specifically, existing research has primarily concentrated on developing service orchestration and task scheduling algorithms, often employing heuristic methods \cite{song2020fast}, reinforcement learning (RL) \cite{song2023reinforcement}, or large language models (LLMs) \cite{tang2025end}. While these approaches have yielded promising results, they frequently consider a limited set of selected factors and AI tasks, which may not capture the full complexity and flexibility of real-world scenarios.

Addressing these limitations necessitates a comprehensive understanding of the myriad factors influencing AI service orchestration in 6G networks. Such an understanding is crucial for developing algorithms that can generalize across various contexts and adapt to dynamic network conditions. Despite the importance of the knowledge, the current literature lacks a systematic compilation and analysis of these factors, creating a gap that hinders the advancement of effective AI service orchestration strategies.

To bridge this gap, our work thoroughly reviews the literature to identify and categorize the factors pertinent to AI service orchestration in 6G networks. Building upon this, we introduce an open-source LLM-assisted toolchain designed to assist AI service providers in constructing comprehensive AI service repositories. This toolchain automates the extraction of relevant factors and facilitates the packaging of AI services, thereby streamlining the deployment process within the 6G AI-RAN environment.

As a proof of concept, we have constructed the Cranfield AI Service repository, incorporating a diverse range of AI services across multiple task categories. This repository leverages open pre-trained models from platforms like HuggingFace, demonstrating the practicality and efficiency of our approach. 

Our specific contributions are:
\begin{itemize}
    \item A comprehensive categorization of AI service orchestration attributes crucial for 6G AI-RAN environments.
    \item An open-sourced, LLM-assisted toolchain for automated AI service repository construction, including model packaging and attribute profiling.
    \item Empirical validation through a case study demonstrating reduced manual coding and highlighting the necessity of infrastructure-aware orchestration.
\end{itemize}

\section{AI Service Orchestration Attributes in 6G}

Effective orchestration of AI services in RAN requires a comprehensive understanding of various attributes that influence service selection, performance, resource allocation, and user satisfaction. 
We summarize these key attributes considered in the literature in Table~\ref{tab:ai_service_attributes} and highlight their significance in the orchestration process.

\begin{table*}[ht]
    \centering
    \caption{Summary of AI service orchestration attributes identified from the literature.}
    \begin{tabular}{p{2cm}|p{3cm}|p{10cm}|p{1.4cm}}
        \toprule
        \textbf{Category} & \textbf{Factor} & \textbf{Definition} & \textbf{References} \\
        
        \midrule
        \multirow{3}{*}{Functionality} 
        & Task Category & High-level tasks the service performs, such as image classification, object detection, depth estimation, text classification etc. & - \\
        \cmidrule{2-4}
        & Task Detail & Detailed description of the expected input and output of the AI service. & - \\
        \cmidrule{2-4}
        & Accuracy & The accuracy of the AI model in relevant benchmarks. & \cite{pan2022joint, liu2023adaptive} \\
        
        \midrule
        \multirow{9}{*}{Resource}
        & CPU Usage & The amount of CPU operations required per AI service execution. &  \cite{zhou2024heterogeneous, pan2022joint, lin2020optimizing, song2023reinforcement} \\
        \cmidrule{2-4}
        & CPU RAM Usage & The volume of CPU memory utilized by the AI service during the execution. & \cite{luo2022container} \\
        \cmidrule{2-4}
        & DEVICE Usage & The amount of accelerator (e.g., GPU) operations required per AI service execution. & \cite{zhou2024heterogeneous, pan2022joint} \\
        \cmidrule{2-4}
        & DEVICE RAM Usage & The volume of accelerator memory required by AI service during the execution & \cite{luo2022container, liang2024resource, zhang2019edge} \\
        \cmidrule{2-4}
        & Disk I/O & The volume of data written to or read from the storage disk per AI service execution. & \cite{luo2022container} \\
        \cmidrule{2-4}
        & Service Storage Size & The disk space required to store the AI model or Docker image of the AI service, which is particularly relevant when deploying large generative AI models on edge nodes. & \cite{liang2024resource, lin2020optimizing} \\
        \cmidrule{2-4}
        & Energy Consumption & The amount of energy consumed by the AI service during both idle and execution states. & \cite{zhou2024heterogeneous, tian2024enabling, heydari2019dynamic, pan2022joint, luo2022container, song2020fast, lin2020optimizing, liu2023adaptive, zhang2019edge} \\
        \cmidrule{2-4}
        & Input Data Size & The data size required per AI service execution. &  \cite{song2020fast, cao2023joint, lin2020optimizing, song2023reinforcement} \\
        \cmidrule{2-4}
        & Output Data Size & The data size generated as output by the AI service & \cite{song2023reinforcement} \\
        
        \midrule
        
        \multirow{3}{*}{Latency}
        & Inference Time & The time taken by the AI service to process one request. & \cite{zhou2024heterogeneous, pan2022joint, luo2022container, cao2023joint, lin2020optimizing, liu2023adaptive, zhang2019edge} \\
        \cmidrule{2-4}
        & Initialization Time & The time the AI service takes to boot up before it is ready for user requests. & \cite{liang2024resource, cao2023joint} \\
        \cmidrule{2-4}
        & Eviction Time & The time taken to undeploy the AI service. & \cite{luo2022container} \\ 
        
        \midrule
        
        Flexibility & Cooperative Inference & Whether the AI service supports cooperative inference across multiple devices or nodes, accepting or outputting intermediate layer features. & \cite{tian2024enabling, zhang2019edge, wang2023social, zeng2024ultra} \\
        
        \midrule
         
        \multirow{2}{*}{Trustworthiness} 
        & Feedback & Number of Likes and Dislikes, as well as general comments by end users. & \cite{ni2024online} \\ 
        \cmidrule{2-4}
        & Explainable AI (XAI) Support & The degree to which the AI service supports integration with XAI methods, facilitating interpretability of its decisions and outputs. & \cite{dutta2023next} \\

        \midrule
        
        \multirow{3}{*}{Billing}
        & Initialization Cost & The amount of credit required to deploy and initialize the AI service & - \\ 
        \cmidrule{2-4}
        & Keep-Alive Cost & The amount of credit cost per time unit to keep the AI service alive listening for user requests. & - \\ 
        \cmidrule{2-4}
        & Execution Cost & The amount of credit cost per AI service execution. & - \\
        
        \bottomrule
    \end{tabular}
    \label{tab:ai_service_attributes}
\end{table*}

\subsection{Functionality Attributes}

\textbf{Task Category and Detail}: Specific functions (e.g., image classification or text analysis) the AI models are trained to perform, the expected input and generated output and exemplar source code on how the AI model can be invoked. Accurately categorizing and describing the tasks ensure that services are precisely matched to the use case requirements especially when queried using semantic searching in vector databases\footnote{https://www.trychroma.com/}\footnote{https://www.mongodb.com/products/platform/atlas-vector-search}.

\textbf{Accuracy}: AI model performance against established benchmarks. High accuracy is crucial for applications where precision is paramount, while resource-constrained scenarios may prioritize resource-light models with lower accuracy due to their reduced resource footprint and faster response time. Worth noting is that the accuracy in operation heavily relies on the discrepancy between the use case's data (in the service request) and the training data of the AI model.

\subsection{Resource Attributes}

\textbf{CPU and DEVICE Usage}: The required computational load, i.e., the amount of CPU and accelerator device (e.g., GPU) operations such as floating-point operations \cite{song2023reinforcement}, ``frequency unit'' \cite{zhou2024heterogeneous} or multiply-and-accumulate operations \cite{pan2022joint}. While determining the accurate number of processor operations might be challenging, these metrics can be replaced by CPU or DEVICE time when profiled on edge or cloud nodes with known CPU/DEVICE frequency.

\textbf{CPU and DEVICE RAM Usage}: The CPU and accelerator device memory needed, which may limit the deployment location of large AI models. Worth noting is that the memory profile also varies due to the size of the request data during service execution.

\textbf{Disk I/O}: The read and write (I/O) data volume on the hard disk, which may incur variable and non-trivial delay when the edge node has limited Disk I/O bandwidth.

\textbf{Service Storage Size}: The storage space required to cache the AI service (e.g., docker image) locally, which may constrain the deployment of large AI models or incur non-trivial download time during initialization. Worth noting is that the size of the AI service image may only reflect the dependencies of the AI services but not the AI model itself as the model parameters can be downloaded and stored in memory during service initialization.

\textbf{Energy Consumption}: The energy consumption when it is pending requests and performing inference. This metric has been a major factor under consideration in many works especially when there are battery-powered computation nodes or the energy is costly.

\textbf{Input and Output Data Size}: The data throughput by AI services, which may affect deployment location or cooperative inference (or semantic communication) arrangements under bandwidth limitations.

\subsection{Latency Attributes}

\textbf{Inference Time}: Either the inference time by the AI model or, more practically, the response time by the AI service to client requests. Like energy consumption, inference time is another major factor considered by the service orchestrators for its direct impact on the Quality-of-Service (QoS). Worth noting is that the inference time varies with the hardware infrastructure ($\propto$ CPU/DEVICE usage divided by CPU/DEVICE frequency) and the current hardware load (when multiple AI services compete for hardware scheduling priority).

\textbf{Initialization and Eviction Time}: The AI service boot-up and termination time. These attributes are often overlooked \cite{liang2024resource} but may pose a significant impact under scenarios with rapidly changing workloads. Similar to inference time, these metrics vary with the infrastructure such as network and memory bandwidth.

\subsection{Flexibility Attributes}

\textbf{Cooperative Inference}: Whether the inference process can be decomposed into subprocesses and distributed across multiple nodes. Such flexibility enables semantic communication which is beneficial upon limited bandwidth or stringent latency requirements.

\subsection{Trustworthiness Attributes}

\textbf{Feedback}: User ratings and comments provide valuable insights into service reliability and effectiveness, which can 1) influence the AI service selection and orchestration strategies, especially for LLM-assisted agents, and 2) inform future improvement and deployment strategies.

\textbf{Explainable AI (XAI) Support}: A list of supported XAI techniques optional as add-ons. Services that offer transparency and explainability in their decision-making processes build user trust and facilitate easier debugging and compliance with regulatory standards.

\subsection{Billing Attributes}

\textbf{Initialization, Keep-Alive, and Execution Costs}: Vital for budgeting and cost management, influencing decisions on service selection and deployment.

By systematically evaluating these attributes, use case innovators and AI service orchestrators can make informed decisions regarding resource allocation, deployment strategies, and service optimization, especially in the LLM-based service provisioning era \cite{tang2025end}.

\section{LLM-assisted Toolchain for AI Service Repositories in AI-RAN}

To facilitate the collection of attributes, we have developed an open-source\footnote{https://github.com/Cranfield-GDP/edge-ai-service-wrapper}, LLM-assisted toolchain designed explicitly for AI service providers in 6G ecosystems. This toolchain facilitates automated packing, functionality validation and attribute profiling of AI services, aiming at reducing the complexity and time required for building deployment-ready service repositories. 

\subsection{Toolchain Overview}

\begin{figure*}
    \centering
    \includegraphics[width=\textwidth]{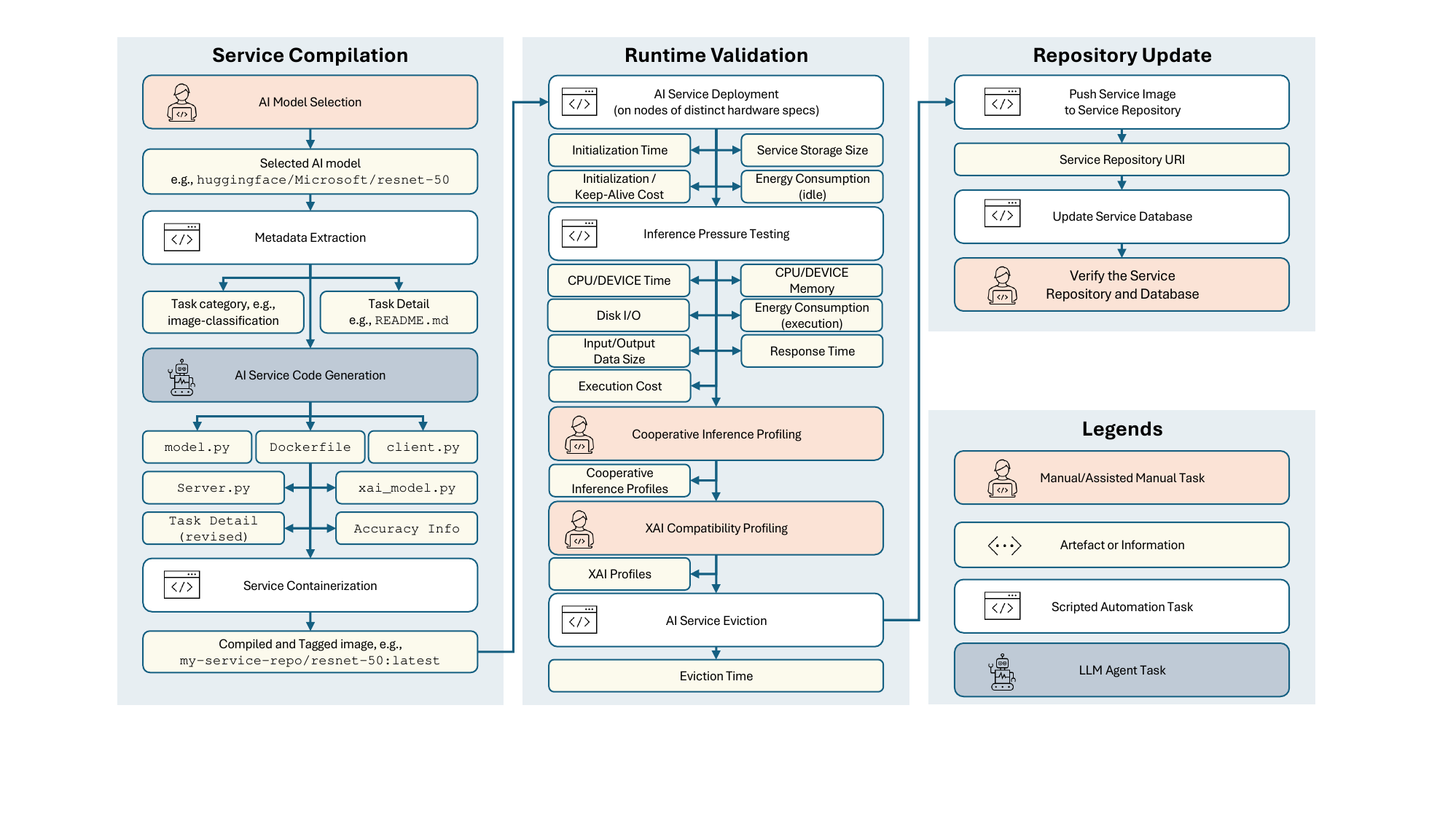}
    \caption{LLM-assisted AI service repository toolchain overview.}
    \label{fig:pipeline_overview}
\end{figure*}

Fig.~\ref{fig:pipeline_overview} illustrates our proposed LLM-assisted toolchain pipeline, comprising three phases:
\textbf{Service Compilation} phase collects and prepares AI service artefacts, including documentation parsing, service code generation, and service image compilation; \textbf{Runtime Validation} phase validates the AI service API endpoints and profiles runtime attributes such as resource usage and latency; and lastly, \textbf{Repository Update} phase updates the AI service repository and database with the generated artefacts and information.

\subsection{Service Compilation Phase}

This phase translates user-selected AI models into deployable AI services with standard service interfaces (e.g., RESTful endpoints):

1) Manual Model Selection: Initially, AI service providers manually select pre-trained AI models of diverse task categories, typically sourced from public repositories such as HuggingFace.

2) Automated Metadata Extraction: 
The model documentation and metadata to identify and categorize task-related information are crawled and parsed automatically, including \textit{task category} and \textit{task detail}.

3) LLM-Assisted Service Code Generation: 
Leveraging large language models (LLMs), the service code is generated automatically based on the AI model documentation and code samples, including:
\begin{itemize}
    \item \texttt{model.py}: RESTful API endpoints encapsulating model inference and runtime profiling endpoints, to be imported by the main AI service application.
    \item \texttt{Dockerfile}: Containerization instructions install dependencies and package the AI service.
    \item \texttt{client.py}: Code examples for requesting AI services.
    \item Revised Task Detail: Enhanced task description for improved semantic searchability in vector databases, tailored particularly for those LLM agent-powered orchestrators \cite{tang2025end}).
    \item Accuracy Information: Performance benchmarks extracted from model documentation if available.
\end{itemize}

To reduce the probability of hallucinations and cost, LLM agents are not required to generate the AI service from scratch; instead, all AI services share a common code base, providing utilities such as the main API server application (\texttt{server.py}), service health checking (\texttt{healthcheck.py}), and explainability add-ons (\texttt{xai\_model.py}) etc.

4) Automated Service Containerization:
The generated service code and dependencies are automatically compiled into a containerized deployment-ready (Docker) image, properly tagged for repository management, such as \texttt{my-service-repo/resnet-50:latest}.

\subsection{Runtime Validation Phase}

Following service compilation, this phase rigorously evaluates the deployability, performance, and operational efficiency of the AI services under realistic network conditions, ensuring that critical runtime attributes are systematically measured, verified, and documented.

5) Automated Service Deployment:
The containerized AI service image is automatically deployed on each edge or cloud node of distinct hardware specs. Upon container instantiation, the following runtime metrics can be captured: \textit{initialization time}, \textit{initialization cost}, \textit{service storage size}, \textit{keep-alive cost} and \textit{idle energy consumption}.

6) Automated Pressure Testing:
Following the deployment, pressure testing is performed using the LLM-generated client scripts to profile service performance and resource utilization at full workload. During this process, detailed runtime attributes can be collected, including: \textit{CPU/DEVICE usage}, \textit{CPU and DEVICE RAM usage}, \textit{disk I/O}, \textit{active energy consumption}, \textit{input and output data size}, \textit{inference time} and \textit{execution cost}.

7) Assisted Manual Cooperative Inference Validation:
This step verifies the AI service’s capability to distribute the inference process across multiple nodes, vital for scenarios demanding semantic communication or real-time latency reduction under constrained network conditions. Results are documented as a list of cooperative inference profiles, each consisting of the start and end layer to be executed on the service side and the corresponding runtime resource metrics discussed above.

8) Assisted Manual XAI Compatibility Validation:
The AI service is then tested against available XAI techniques, which is essential for debugging, user trust and regulatory compliance. The results are documented as a list of XAI technique profiles, each consisting of the name of the XAI technique as well as the corresponding runtime resource metrics discussed above.

9) Automated Service Eviction:
Finally, the AI service is un-deployed from the hosting nodes and the attribute \textit{eviction time} is measured.

\subsection{Repository Update Phase}
10) After comprehensive validation, the AI service images are pushed to a service repository, each identified by a unique URI. Then, the service database is populated with detailed metadata including all essential attributes collected during the service compilation and runtime validation phase. To maintain quality and trust, the AI service provider can conduct a final verification step, ensuring that the saved service knowledge is accurate, complete, and consistent.

\section{Case Study: Cranfield AI Service Repository}

To validate the practicality and effectiveness of our proposed toolchain, we implemented a proof-of-concept AI Service repository hosted at Cranfield University\footnote{https://hub.docker.com/repositories/cranfield6g}. This repository currently integrates more than 30 AI services performing 12 different tasks using publicly available pre-trained models from HuggingFace. New AI services can be added within minutes.

\begin{figure}
    \centering
    \includegraphics[width=\columnwidth]{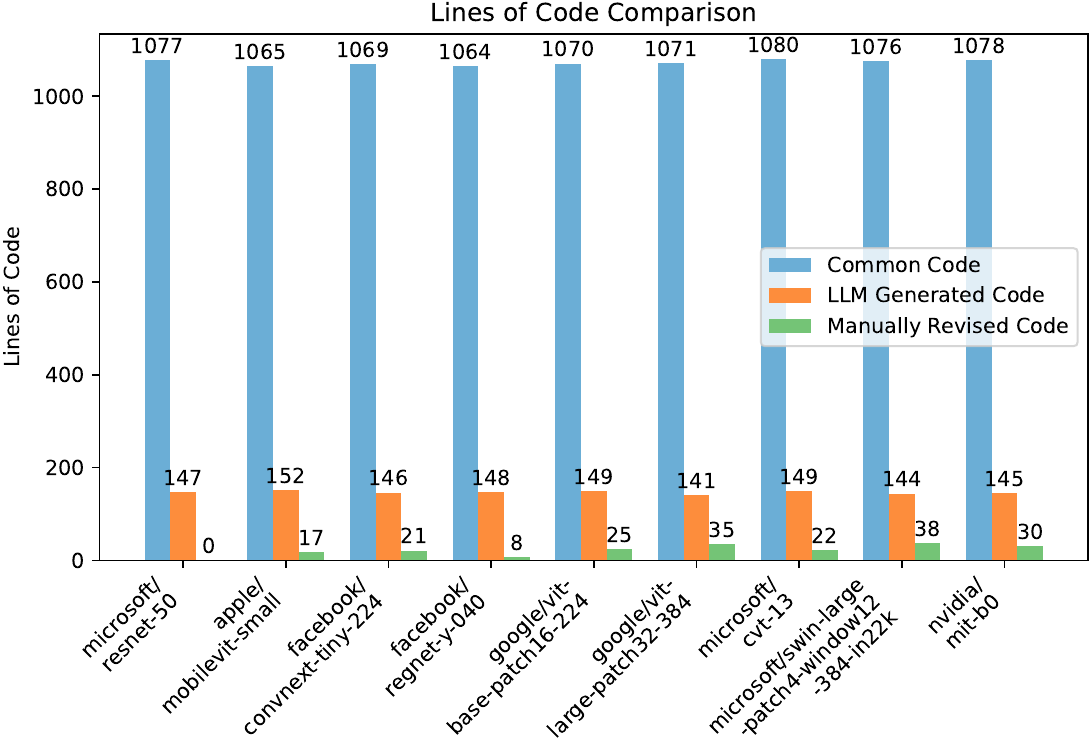}
    \caption{Lines of code comparison for the first nine AI services (for image classification) in Cranfield AI Service Repository. The pipeline has been developed using the model \textit{microsoft/resnet-50} as a reference, hence zero manual revisions counted for the model.}
    \label{fig:line-of-code-comparison}
\end{figure}

\subsection{AI Service Compilation Automation}

Fig.~\ref{fig:line-of-code-comparison} compares the lines of codes on the common code base (which varies mainly due to formatting differences), LLM agent-generated AI model-specific code base and human developer adjustments. 

\textbf{Observation} Thanks to the mature software engineering practices, a significant portion of the code base can be shared across AI services hosting different AI models. LLM-based agents are only required to fill those small model-specific pieces which effectively minimizes the cost and the probability of hallucinations. Manual code revisions are still required to deal with unexpected issues such as zero-day dependency breaks \footnote{https://github.com/huggingface/transformers/issues/37326} (i.e., compilations of the same code base may succeed at first and fail later) or to select the best target neural network layer for XAI or cooperative inference\footnote{https://jacobgil.github.io/pytorch-gradcam-book/HuggingFace.html}. After all, XAI techniques are designed to produce stakeholder-centred (e.g., AI service provider, AI service user and legal auditor) explanations and thus the stakeholders shall remain in the loop.

\begin{tcolorbox}[colback=cyan!5!white, colframe=cyan!40!black, title=Takeaway 1: Challenge of Full Automation] 
Manual effort (or stakeholder involvement in general) can be minimized utilizing such LLM-powered toolchains but may never be eliminated, because 1) most AI models are manually designed and trained, 2) most AI service dependencies are manually maintained and 3) trustworthiness is stakeholder-specific.
\end{tcolorbox}

\subsection{AI Service Resource Profiling}

\begin{figure*}[ht]
    \centering
    \includegraphics[width=\textwidth]{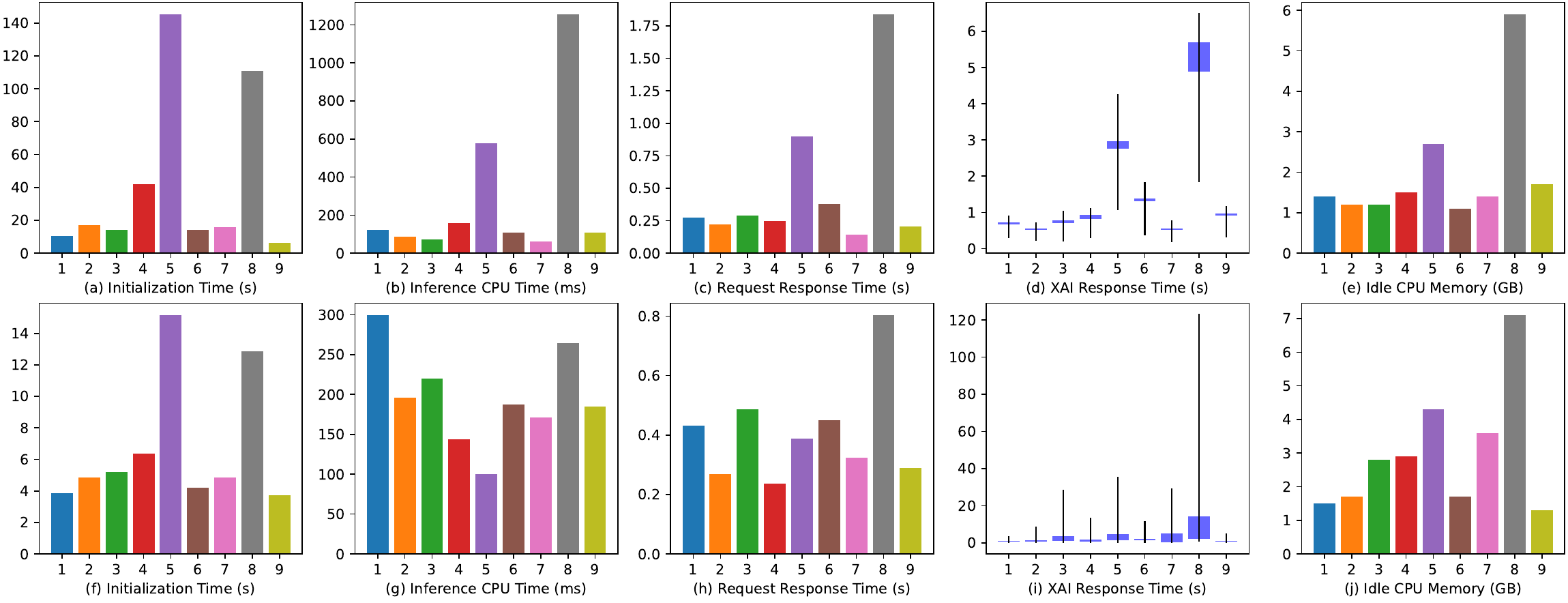}
    \caption{Resource profile results of the first nine AI services. (a-e) are profiled on a CPU-only laptop connected to WIFI and (f-j) on an RTX A4000 GPU-powered workstation with a faster ethernet connection. The candlestick plots for XAI response time are generated from all compatible XAI techniques for each AI service.}
    \label{fig:resource-profile-results}
\end{figure*}

From the runtime resource profiling results in Fig~\ref{fig:resource-profile-results}, we learn that the edge nodes of different hardware and network infrastructures can produce significantly different AI service runtime profiles. Worth noting is that two XAI techniques (ScoreCAM and AblationCAM) are not even supported due to memory shortage on the laptop testbed.

\begin{tcolorbox}[colback=cyan!5!white, colframe=cyan!40!black, title=Takeaway 2: Infra-specific AI Service Profile] 
Most literature assumes the latency, resource profile and the task (e.g., compatible XAI techniques) of the orchestrated AI services are infrastructure-agnostic while in reality, they vary with different hardware specs and real-time hardware load. Lack of such consideration renders many proposed service orchestrators ineffective in real settings.
\end{tcolorbox}

\section{Conclusions and Future Works}
\textbf{Conclusions} We presented an LLM-assisted toolchain that significantly automates AI service packaging and profiling for effective orchestration in 6G AI-RAN environments. The empirical case study on building the Cranfield AI Service Repository illustrated substantial manual effort reductions and underscored the importance of infrastructure-specific profiling. We hope our work enables more practical, efficient service orchestration solutions in 6G AI-RAN.

\textbf{Future Works} we cordially invite the community to join our endeavour in making the toolchain truly beneficial to the telecom ecosystem. Specifically, we believe the toolchain could benefit significantly in 1) supporting metrics or techniques to evaluate or enhance other essential trustworthy metrics than explainability, such as safety, security, robustness, privacy, and bias; 2) supporting AI training services, particularly federated learning which is widely adopted in 6G RAN and use cases; 3) supporting multi-AI pipelines incorporating multiple AI models for complex use cases; 4) supporting AI service transfer learning techniques to adapt pre-trained AI models to new use cases; and 5) supporting AI model quantization for different hardware and network infrastructures.

\bibliographystyle{IEEEtran}
\bibliography{references}

\end{document}